\documentclass[10pt,journal,compsoc]{IEEEtran}
\ifCLASSOPTIONcompsoc
  \usepackage[nocompress]{cite}
\else
  \usepackage{cite}
\fi
\ifCLASSINFOpdf
  \usepackage[pdftex]{graphicx}
\else
  \usepackage[dvips]{graphicx}
\fi
\usepackage{amsmath}
\usepackage{amsfonts}
\usepackage{bm}
\usepackage{booktabs}

\usepackage{wrapfig}
\usepackage{float}
\usepackage{color}
\usepackage[dvipsnames]{xcolor}
\usepackage{colortbl}

\newcommand{\myColor}{black}
\newcommand{\maybeColor}[1]{\textcolor{\myColor}{#1}}
\newcommand{\best}{\cellcolor{tablered}}
\newcommand{\sbest}{\cellcolor{orange}}
\newcommand{\tbest}{\cellcolor{yellow}}
\definecolor{yellow}{rgb}{1, 1, 0.7}
\definecolor{orange}{rgb}{1, 0.85, 0.7}
\definecolor{tablered}{rgb}{1, 0.7, 0.7}

\usepackage{svg}

\begin{document}
%
\title{2DGH: 2D Gaussian-Hermite Splatting for High-quality Rendering and Better Geometry Features}
%
%
%
%

\author{Ruihan~Yu*,
        Tianyu~Huang*,
        Jingwang~Ling and Feng~Xu$^\dagger$
\IEEEcompsocitemizethanks{\IEEEcompsocthanksitem R. Yu is with Department of Physics, Tsinghua University. 
\IEEEcompsocthanksitem J. Ling, T. Huang and F. Xu are with School of software and BNRist, Tsinghua University.}

\thanks{*Equal contribution.}
\thanks{$^\dagger$Corresponding author. Email: {\texttt{xufeng2003@gmail.com}}.}
\thanks{\textcopyright~2025 IEEE. Personal use of this material is permitted. Permission from IEEE must be obtained for all other uses, in any current or future media, including reprinting/republishing this material for advertising or promotional purposes, creating new collective works, for resale or redistribution to servers or lists, or reuse of any copyrighted component of this work in other works. DOI: 10.1109/TVCG.2025.3622157.}
}
%
%

\markboth{IEEE Transactions on Visualization and Computer Graphics, 2025}%
{Yu \MakeLowercase{\textit{et al.}}: 2DGH: 2D Gaussian-Hermite Splatting for High-quality Rendering and Better
Geometry Features}
%



\IEEEtitleabstractindextext{%
\begin{abstract}
 2D Gaussian Splatting has recently emerged as a significant method in 3D reconstruction, enabling novel view synthesis and geometry reconstruction simultaneously. While the well-known Gaussian kernel is broadly used, its lack of anisotropy and deformation ability leads to dim and vague edges at object silhouettes, limiting the reconstruction quality of current Gaussian splatting methods. To enhance the representation power, we draw inspiration from quantum physics and propose to use the Gaussian-Hermite kernel as the new primitive in Gaussian splatting. The new kernel takes a unified mathematical form and extends the Gaussian function, which serves as the zero-rank special case in the updated general formulation. Our experiments demonstrate that the proposed Gaussian-Hermite kernel achieves improved performance over traditional Gaussian Splatting kernels on both geometry reconstruction and novel-view synthesis tasks. Specifically, on the DTU dataset, our method yields more accurate geometry reconstruction, while on datasets such as MipNeRF360 and our customized Detail dataset, it achieves better results in novel-view synthesis. These results highlight the potential of the Gaussian-Hermite kernel for high-quality 3D reconstruction and rendering.
\end{abstract}

}

\maketitle

\IEEEdisplaynontitleabstractindextext

%
\IEEEpeerreviewmaketitle


%
%
%
%

\begin{figure*}[t]
	\centering
	\includegraphics[width=.99\linewidth]{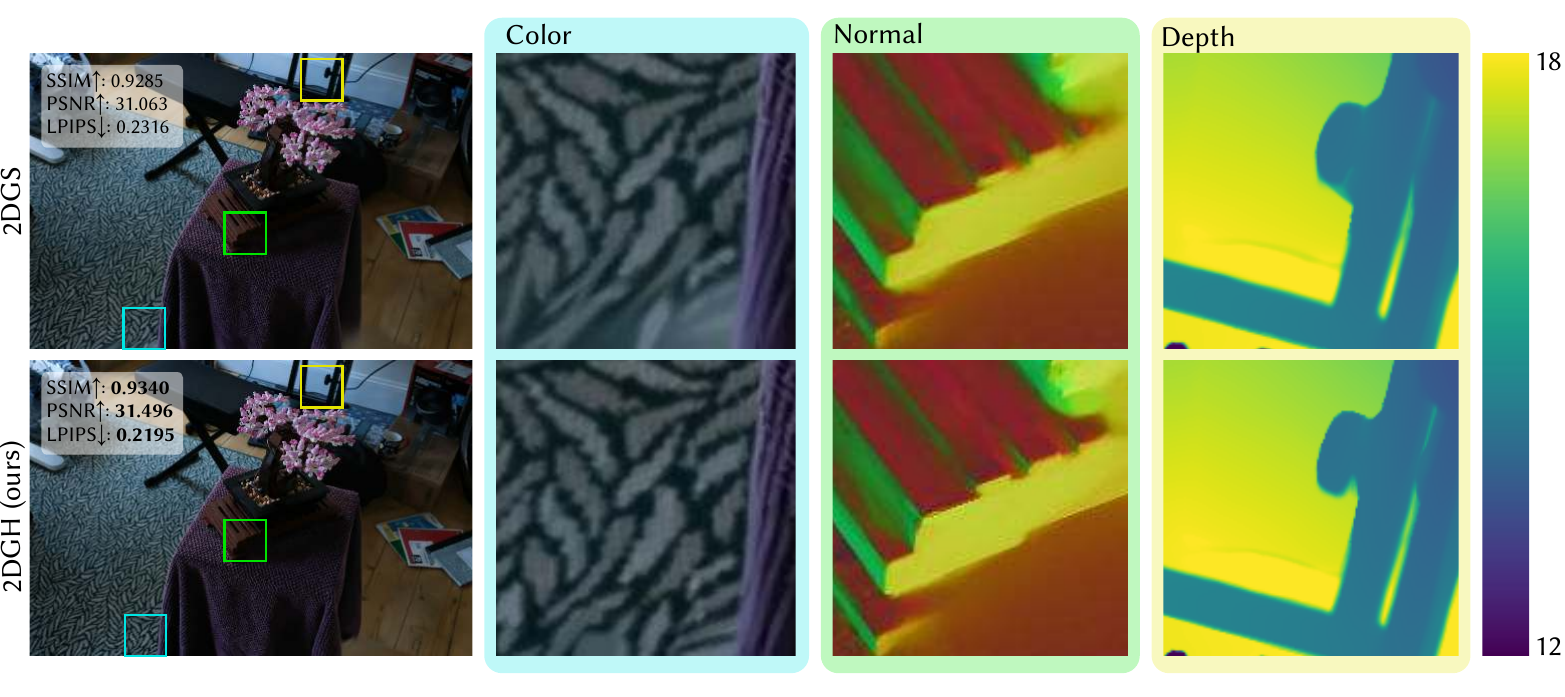}
	\caption{By modulating the Gaussian functions with Hermite series, the resulting Gaussian-Hermites exhibit a stronger representational capacity than original 2D Gaussians, particularly excelling in the reconstruction of fine complex structures and sharp discontinuous edges.
    Compared to the current state-of-the-art 2D Gaussian Splatting (2DGS)~\cite{huang20242dgs}, our proposed method, 2D Gaussian-Hermite Splatting (2DGH), demonstrates superior novel view synthesis performance in highly complex scenes while achieving comparable or even better geometric reconstruction quality under the same number of Gaussians. %
}
\label{fig:teaser}
\end{figure*}

\section{Introduction}

\IEEEPARstart{N}{ovel-view} synthesis (NVS) and 3D surface reconstruction remain fundamental challenges in computer vision and graphics. These tasks are central to the broader goal of inverse rendering, which seeks to recover scene properties—such as geometry, appearance, and lighting—from observed images. A prominent line of work in this area is Neural Radiance Fields (NeRF)~\cite{mildenhall2020nerf}, which models volumetric scenes using neural networks and has demonstrated impressive results in photorealistic view synthesis. However, despite their success, NeRF-based methods often suffer from high computational costs and slow rendering speeds, which hinder their deployment in real-time applications.

Recently, 3D Gaussian Splatting (3DGS)~\cite{kerbl2023gaussian} achieves an optimal balance between real-time performance and high-fidelity rendering, addressing long-standing limitations in inverse rendering pipelines.
Although the volumetric nature of the 3D Gaussian kernel enables various extensions such as global illumination~\cite{zhou2024ugp}, dynamic scenes~\cite{yang2023deformable3dgs, huang2023sc, yan2024dynamic}, and anti-aliasing~\cite{yu2023mipsplatting, yan2024multiscale, song2024sa, liang2024analytic}, the kernel essentially lacks a good definition of surfaces and fails to extract high-quality meshes.

To achieve better surface reconstruction, subsequent research efforts, such as SuGAR~\cite{guedon2023sugar}, 2DGS~\cite{huang20242dgs}, and Gaussian Surfels~\cite{dai2024gaussiansurfels} 
strive to modify the kernels to make them behave more like surfaces.
From the perspective of mesh-based rendering, these research efforts can be seen as bridging the gap between splatting primitives and traditional mesh facets, effectively pushing volumetric representations like 3DGS toward mesh-based representations with more clearly defined surfaces.

However, there remains another gap between the shape primitives of polygon meshes and Gaussian splitting. 
Unlike polygons, which have clear edges, elliptical Gaussian kernels lack the flexibility to be deformed in a way that can express sharp boundaries and complicated structures effectively. Enhancing the representation power of the shape primitives holds potential in improving the reconstruction quality of current Gaussian splatting methods.

On the other hand, the Gaussian function has a well-established history in various fields, such as statistics~\cite{liu1994note, challa2000nonlinear}, physics~\cite{siegman1973hermite, dekker1981classical, kimel1993relations}, and electronics~\cite{yao2016patch,li2012line}, holding an irreplaceable role in signal analysis~\cite{kong2008analytic, yang2011image}. 
The wave function in quantum physics shares conceptual similarities with Gaussian Splatting, both of which describe the spatially-varying distribution of matter or particles using Gaussian functions. 
Furthermore, quantum physics indicates that beyond the standard Gaussian, there exists a family of higher-rank Gaussian functions, such as Gaussian-Hermite polynomials~\cite{Sakurai_Napolitano_2020}. 
These Gaussian-Hermite polynomials serve as solutions to the quantum harmonic oscillation equation, a fundamental model in quantum physics. 
Also, they are used to address the interaction between photons and electrons, which is directly relevant to the underlying physics of volume scattering in physics-based rendering~\cite{chandrasekhar2013radiative}.
Consequently, one can expect that the Gaussian-Hermite polynomials, which model electron orbital distributions in quantum physics, could potentially be adopted to model the opacity distribution in Gaussian splitting. 
The opacity distribution can be regarded as a macroscopic perspective of the electron orbital distribution, as both reflect the fundamental forms that compose the matter we observe.
 
Inspired by research on Gaussian-based functions in quantum physics, we propose using a unified representation, Gaussian-Hermite polynomials as the kernels for Gaussian Splatting. 
With higher-rank terms, these new kernels can deform beyond the elliptical shape and better express sharp boundaries. 
However, naive addition of high-rank functions can result in invalid opacity values. 
Therefore, we propose techniques to handle large and negative coefficients of the high-rank Gaussian-Hermite function, ensuring the resulting opacity falls within the valid range of $[0,1]$.
We compare our proposed kernel with previous ones on Gaussian Splatting tasks, and our experiments show that the Gaussian-Hermite polynomial kernel achieves state-of-the-art performance in terms of NVS and geometry reconstruction quality compared to previous Gaussian kernels.
Notice that modifying the shape primitive is independent of and compatible with recent advancements in other aspects of Gaussian splatting.

In summary, we make the following contributions: 
\begin{itemize}
    \item We first introduce a family of Gaussian-Hermite kernels with higher representation power in Gaussian Splatting, and prove that the original Gaussian kernels are the zero-rank case of the new formulation.
    \item To enable the use of Gaussian-Hermite polynomials in opacity modeling, we propose a new activation function that can handle high-order coefficients in the alpha-blending process.
    \item We introduce a synthetic dataset comprising objects with complex geometries and material properties, as a benchmark for evaluating the fitting capabilities of 3D representations.
    \item We conduct experiments to compare our proposed kernels with previous ones in Gaussian Splatting, showing that our method improves reconstruction around shape boundaries and achieves state-of-the-art performance in surface reconstruction and novel-view synthesis.
\end{itemize}

\section{Related Work}
\label{sec:related}

\subsection{Novel View Synthesis}
Performing novel view synthesis (NVS) from a set of input images has been central to computer graphics and vision research. Neural Radiance Fields (NeRF)~\cite{mildenhall2020nerf} utilizes neural networks to model the radiance distribution in space, thereby enabling NVS. Following NeRF, a significant number of NeRF-based works have notably expanded its capabilities, such as alleviating the aliasing issues~\cite{barron2021mip, barron2022mip360, hu2023trimip}, extending NeRF to unbounded scenes~\cite{zhang2020nerf++, barron2022mip360}, and improving rendering~\cite{yu2021plenoctrees, reiser2021kilonerf, hedman2021baking, reiser2023merf, yariv2023bakedsdf, chen2023mobilenerf} and training~\cite{liu2020neural, chen2022tensorf, sun2022directvoxel, mueller2022instant, yu2022plenoxels, barron2023zip, chen2023neurbf} efficiency. 

The emergence of 3D Gaussian Splatting (3DGS)~\cite{kerbl2023gaussian} has allowed NVS with real-time framerate, with further extensions~\cite{yu2023mipsplatting, qian2023gaussian, zielonka2023drivable, xie2023physgaussian, jiang2023relightable, zhou2024ugp, guedon2023sugar, huang2023sc, yu2024gof, yang2023deformable3dgs, yan2024multiscale, song2024sa, liang2024analytic}. Among the enhancements to 3DGS, GES~\cite{hamdi2024ges} reformulates the mathematical representation of 3D Gaussians by introducing the Generalized Exponential Function (GEF) to represent 3D scenes, demonstrating a stronger scene representation capability compared to the original 3D Gaussians. 

Recently, 2D Gaussian Splatting (2DGS)~\cite{huang20242dgs} has gained attention as an innovative approach that simplifies 3D scene representation by reducing volumetric data into 2D oriented Gaussian disks. 2DGS demonstrates superior geometric reconstruction performance compared to 3DGS without sacrificing efficiency; however, it exhibits a decline in both qualitative and quantitative metrics for NVS. 

Our work introduces a new mathematical representation for 2D Gaussians involving Hermite series, which enhances the scene representation capability of 2DGS.

\subsection{Multi-view 3D Reconstruction}
Multi-view 3D reconstruction has seen substantial advancements with traditional methods like Multi-view Stereo (MVS)~\cite{schonberger2016pixelwise, yao2018mvsnet, yu2020fastmvsnet}, but has been further revolutionized by neural approaches. Neural approaches such as Occupancy Networks~\cite{mescheder2019occupancy} and Neural Implicit Surfaces~\cite{niemeyer2020differentiable, yariv2020multiview} use multi-layer perceptrons (MLPs) to represent 3D geometry. Subsequent works~\cite{wang2021neus, oechsle2021unisurf, yariv2021volume} employs volume rendering techniques to render implicit surfaces. More recent innovations include scalable approaches~\cite{li2023neuralangelo, yu2022sdfstudio, yu2022monosdf} and methods that target high efficiency in complex scenes~\cite{yu2021plenoctrees, chen2022tensorf, wang2023neus2}. Methods based on Neural Implicit Surfaces and volume rendering are not highly efficient. 3DGS~\cite{kerbl2023gaussian} employs an explicit scene representation, significantly improving training and rendering speed. However, the lack of well-defined boundaries results in degraded geometric reconstruction quality. To address this issue, various methods, including SuGaR~\cite{guedon2023sugar}, GOF~\cite{yu2024gof}, Gaussian Surfels~\cite{dai2024gaussiansurfels}, and 2DGS~\cite{huang20242dgs}, have been proposed. The most recent and well-known 2DGS has demonstrated state-of-the-art geometric reconstruction performance. Building on 2DGS, our approach introduces a new Gaussian mathematical formulation, improving the reconstruction of fine structures and discontinuous surface connections.

\section{Method}
\label{sec:methods}
\subsection{Preliminaries}

3D Gaussian Splatting (3DGS)~\cite{kerbl2023gaussian} proposes to represent 3D scenes with
many translucent 3D Gaussian ellipsoids and render images through rasterization. Specifically, 3DGS parameterizes Gaussian primitives via mean position $\boldsymbol{\mu}_k$, opacity $\alpha$ and covariance matrix $\Sigma$: 
\begin{equation}
  G(\boldsymbol{x}, \alpha, \Sigma) = \alpha \exp\left(-\frac{1}{2}(\boldsymbol{x}-\boldsymbol{\mu}_k)^\top \Sigma^{-1} (\boldsymbol{x}-\boldsymbol{\mu}_k)\right),
  \label{eq:3dgs}
\end{equation}
where $\boldsymbol{x}$ is a position in 3D world space and the covariance matrix is factorized into a rotation matrix $\boldsymbol{R}$ and a scaling matrix $\boldsymbol{S}$ :
\begin{equation}
    \Sigma = \boldsymbol{R} \boldsymbol{S}\boldsymbol{S}^\top \boldsymbol{R}^\top.
  \label{eq:3dgs_cov}
\end{equation}

However, 3DGS suffers from poor geometry reconstruction performance due to view inconsistency. 2DGS~\cite{huang20242dgs} proposes that diminishing one scaling to zero and making primitive more like surface can improve view consistency. In 2DGS, Gaussians can be represented by:
\begin{equation}
    G(\boldsymbol{x}) = \exp\left(-\frac{u(\boldsymbol{x})^{2} + v(\boldsymbol{x})^{2}}{2}\right),
  \label{eq:2dgs}
\end{equation}
where $u(\boldsymbol{x})$ and $v(\boldsymbol{x})$ are local UV space coordinates. 

In the rasterization process, 3DGS will project the Gaussian primitives onto a 2D manifold with EWA approximation~\cite{zwicker2001ewa}. Previous work~\cite{zwicker2004perspective} has demonstrated that this projection is reliable only at the center of the Gaussian function, with the approximation becoming less precise as the query position moves away from the center. 2DGS~\cite{huang20242dgs} shows that it will lead to unstable optimization during differentiable rendering, thus introducing the Ray-Splat Intersection~\cite{sigg2006gpu} to address this issue.

Given central position $\boldsymbol{p}_{k}$, a scaling vector $\boldsymbol{S} = (s_{u}, s_{v})$ that controls the covariance of a 2D
Gaussian and a $3 \times 3$ rotation matrix $\boldsymbol{R} = [\boldsymbol{t}_{u}, \boldsymbol{t}_{v}, \boldsymbol{t}_{w}]$ that controls 2D Gaussian orientation, the transformation between UV space and world space can be written as the following:
\begin{equation}
    \boldsymbol{H} = \begin{bmatrix}
        s_{u}\boldsymbol{t}_{u} & s_{v}\boldsymbol{t}_{v} & \boldsymbol{0} & \boldsymbol{p} \\
        0 & 0 & 0 & 1
    \end{bmatrix} = \begin{bmatrix}
        \boldsymbol{RS} & \boldsymbol{p}_{k} \\
        \boldsymbol{0} & 1 
    \end{bmatrix}.
  \label{eq:2dgs_param}
\end{equation}

Assuming $\boldsymbol{W}$ is the transformation matrix from world space to screen space, a homogeneous ray started from the camera and passing through pixel (x, y) can be obtained by:
\begin{equation}
    \boldsymbol{x} = (xz, yz, z, 1)^\top = \boldsymbol{W}\boldsymbol{H}(u, v,1,1)^\top,
  \label{eq:2d_projection}
\end{equation}
where z represents intersection depth. In the rasterization, we input pixel coordinate $(x, y)$ and inquiry intersection in Gaussian's local coordinate. To achieve that, we need to obtain the inverse transformation of the projection Eq. (\ref{eq:2d_projection}). The intersection depth z is constrained by the view-consistent 2D Gaussian. Therefore, by solving this constraint equation, we can get the final result, as detailed in~\cite{huang20242dgs}: 
\begin{equation}
    u(\boldsymbol{x}) = \frac{\boldsymbol{h}_{u}^{2}\boldsymbol{h}_{v}^{4} - \boldsymbol{h}_{u}^{4}\boldsymbol{h}_{v}^{2}}{\boldsymbol{h}_{u}^{1}\boldsymbol{h}_{v}^{2} - \boldsymbol{h}_{u}^{2}\boldsymbol{h}_{v}^{1}},\quad v(\boldsymbol{x}) = \frac{\boldsymbol{h}_{u}^{4}\boldsymbol{h}_{v}^{1} - \boldsymbol{h}_{u}^{1}\boldsymbol{h}_{v}^{4}}{\boldsymbol{h}_{u}^{1}\boldsymbol{h}_{v}^{2} - \boldsymbol{h}_{u}^{2}\boldsymbol{h}_{v}^{1}} ,
  \label{eq:2d_projection_inverse}
\end{equation}
\begin{equation}
    \boldsymbol{h}_{u} = (\boldsymbol{WH})^\top(-1,0,0,x)^\top ,\quad\boldsymbol{h}_{u} = (\boldsymbol{WH})^\top(0,-1,0,y)^\top,
  \label{eq:2d_projection_inverse_2}
\end{equation}
where $(x,y)$ is the pixel coordinate and $\boldsymbol{h}_{u}^{i}$, $\boldsymbol{h}_{v}^{i}$ represent the i-th parameter of the vector.

Unlike 3DGS, 2DGS maintains a strict one-to-one mapping between points on the 2D Gaussians and pixel coordinates in screen space. This leads to another unexpected benefit---the ability to freely modify the expression of the Gaussian kernel. This is because the transformation from the local coordinate system to the screen coordinate system is a reversible transformation in 2DGS, whereas 3DGS utilizes an irreversible non-affine transformation.

For our method, this reversibility property provides a great deal of convenience.

\subsection{Gaussian-Hermite Splatting Kernel}

Gaussian-Hermite (GH) polynomials are the product of a Gaussian function and Hermite polynomials. In mathematics, Hermite polynomials frequently appear where a Gaussian distribution is utilized. This family of polynomials has applications in various fields,  such as the eigenmode of laser beam~\cite{siegman1973hermite}, the solution of Appell's equation of motion in classical mechanics~\cite{desloge1988gibbs}, Edgeworth series in probability theory~\cite{challa2000nonlinear} and quantum harmonic oscillator in quantum mechanics~\cite{Sakurai_Napolitano_2020}.

The $n$-th rank Hermite polynomial, denoted as $H_n(x)$, is defined by the Rodrigues formula~\cite{davis2024general}:
\begin{equation}
    H_{n}(x) = \frac{(-1)^{n}}{\omega(x)}\frac{\partial^{n}}{\partial x^{n}} \omega(x),
  \label{eq:Hermite_definition}
\end{equation}
\begin{equation}
     \omega(x) = \frac{1}{\sqrt{2\pi}}\exp(-\frac{x^{2}}{2}).
  \label{eq:Hermite_definition_2}
\end{equation}

\begin{figure}[t]
\centering
\includegraphics[width=\linewidth]{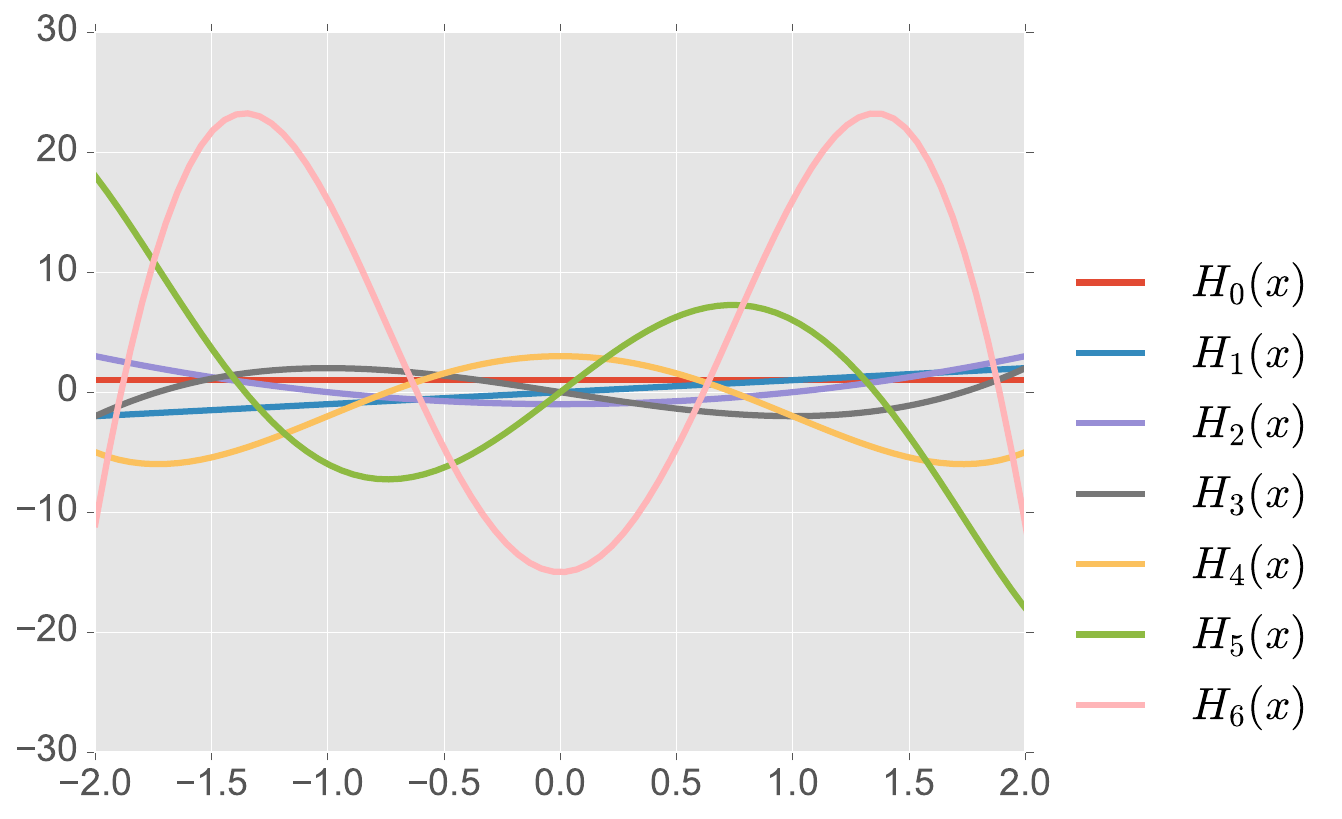} 

\vspace{-0.5 em}
\caption{
\textbf{Hermite Polynomials.}
The figure presents Hermite polynomials $H_n(x)$ for $n = 0, 1, 2, \dots, 6$. These orthogonal polynomials exhibit varying degrees of oscillatory behavior as $n$ increases.
}
\label{fig:hermite_poly}
\end{figure}

The following are first 9 rank Hermite polynomials from $H_0(x)$ to $H_8(x)$:
\begin{align*}
    H_{0}(x) &= 1 , \\
    H_{1}(x) &= x , \\
    H_{2}(x) &= x^{2}-1 , \\
    H_{3}(x) &= x^{3}-3x , \\
    H_{4}(x) &= x^{4}-6x^{2}+3 , \\
    H_{5}(x) &= x^{5}-10x^{3}+15x , \\
    H_{6}(x) &= x^{6}-15x^{4}+45x^{2}-15 , \\
    H_{7}(x) &= x^{7}-21x^{5}+105x^{3}-105x , \\
    H_{8}(x) &= x^{8}-28x^{6}+210x^{4}-420x^{2}+105 .
\end{align*}
$H_0(x)$ to $H_6(x)$ are also visualized in Fig. \ref{fig:hermite_poly}.

 Actually, Hermite polynomials form a set of complete orthogonal basis~\cite{johnston2014weighted}. A simple proof would be writing monomials as a finite linear combination of Hermite polynomials:
\begin{equation}
    x^{n} = a_{0}H_{0}(x) + a_{1}H_{1}(x) + ... + a_{n}H_{n}(x),
  \label{eq:linear_comb}
\end{equation}
where $a_{0}$, $a_{1}$, ..., $a_{n}$ stand for decomposition coefficients.
It is well known that monomials are a complete basis in $L^2(\mathbb{R})$ space, which means they are square-integrable functions. 

Therefore, any function belonging to the $L^2(\mathbb{R})$ space (most cases in the real world) can be written as:
\begin{equation}
    f(x) = \sum_{n=0}^{\infty}a_{n}H_{n}(x).
  \label{eq:func_hermite_expr}
\end{equation}

\noindent
After multiplying a Gaussian function, we can still prove the GH polynomials form a complete orthogonal basis in the weighted Hilbert space $L^2(\mathbb{R})$, as demonstrated in the Appendix.
Consequently, any function $f \in L^2(\mathbb{R})$ admits a spectral expansion of the form:
\begin{equation}
    f(x) = \sum_{n=0}^{\infty}a_{n}GH_{n}(x),
  \label{eq:func_gh_expr}
\end{equation}

\begin{equation}
    GH_{n}(x) = e^{-\frac{1}{2}x^{2}}H_{n}(x).
  \label{eq:1dgh}
\end{equation}

\begin{figure}[t]
\centering
\includegraphics[width=\linewidth]{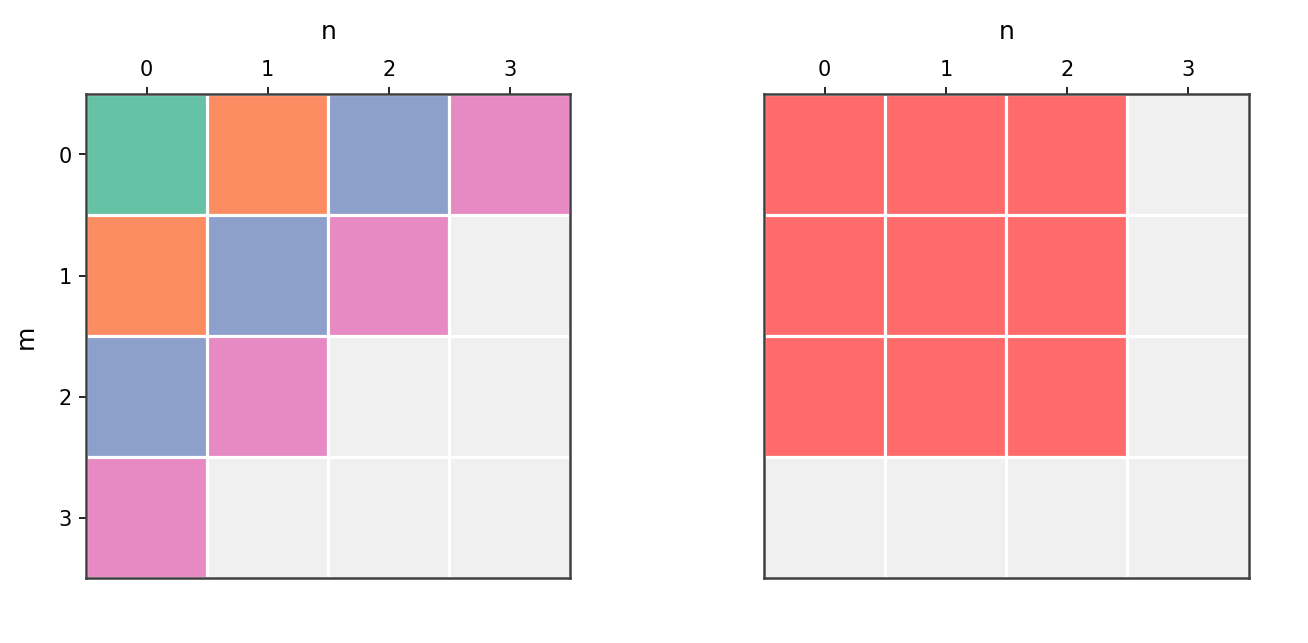} 

\vspace{-0.5 em}
\caption{
\textbf{Visualization comparison between pyramidal and square coefficient arrangements.}
Our analysis demonstrates that the 9-coefficient configuration in the right panel (square grid) yields better performance compared to the 10-coefficient layout in the left panel (pyramidal structure).
}
\label{fig:coeff_pyramid}
\end{figure}

For 2D Gaussian Splatting, we propose changing Gaussian kernel into GH kernel:
\begin{equation}
    \begin{split}
    f(\boldsymbol{x}) &= \exp\left(-\frac{u(\boldsymbol{x})^{2} + v(\boldsymbol{x})^{2}}{2}\right) \times\\
    &\qquad \left(\sum_{m=0}^{M}\sum_{n=0}^{N}c_{mn}H_{m}(u(\boldsymbol{x}))H_{n}(v(\boldsymbol{x}))\right),
      \label{eq:gh_kernel}
  \end{split}
\end{equation}
where $c_{mn}$ are optimizable coefficients, $N$ and $M$ stand for maximum rank numbers in orthogonal directions. People usually control the total rank $(N+M)$, which conventionally enforces a pyramidal coefficient arrangement (Fig. \ref{fig:coeff_pyramid} left, common in physical and math applications), like SH coefficients used by Gaussian Splatting \cite{kerbl2023gaussian, huang20242dgs}. However, we find it's more efficient to use a square arrangement with $N=M$ given a similar amount of coefficients (Fig. \ref{fig:coeff_pyramid} right). For our method, we set $N=M=3$. Different from GES~\cite{hamdi2024ges}  with only one optmizable parameter, our method can be easily scaled up to higher ranks to support stronger fit capacity for one primitive by Hermite recursion:
\begin{equation}
    \begin{split}
    H_{n+1}(x) &= xH_{n}(x) - nH_{n-1}(x).
      \label{eq:gh_recursion}
  \end{split}
\end{equation}

\begin{figure}[t!]
\centering
\includegraphics[width=\linewidth]{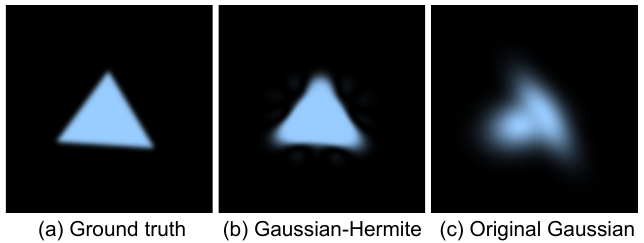}
\vspace{-1.5 em}
\caption{
\textbf{Toy experiment.}
We use 2 original Gaussians or 2 Gaussian-Hermites to fit the triangle in (a). In (b), two 5-rank (M=N=5) Gaussian-Hermites more sharply captures the edge features of the triangle. However, in (c), the result produced by fitting with 2 original Gaussians only vaguely represents the general shape of the triangle and fails to clearly capture its edge features.
}  
\label{fig:toy}
\end{figure}

\begin{figure}[t]
\centering
\includegraphics[width=\linewidth]{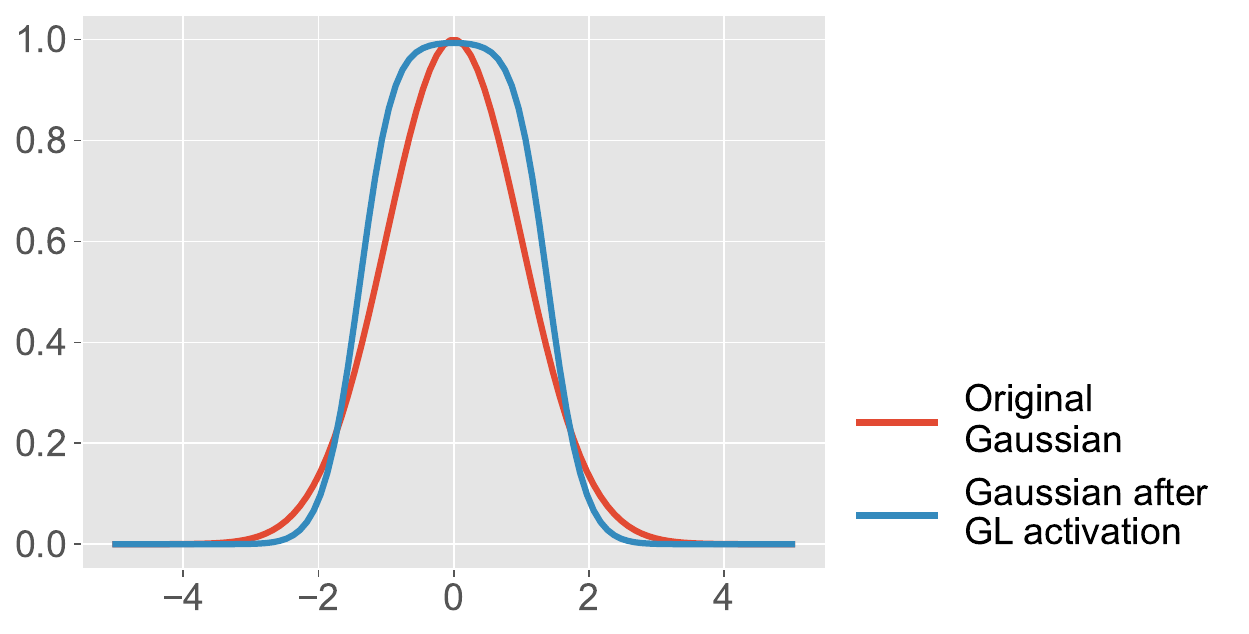} 

\vspace{-0.5 em}
\caption{
\textbf{Visualization comparing the original Gaussian with the Gaussian after applying GL activation function.}
Gaussian after GL activation is smoother near the mean and exhibits a steeper decline as it moves away from the mean.
}
\label{fig:new_ac}
\end{figure}

In Fig. \ref{fig:toy}, we show how 2D Gaussians can fit a mesh triangle on a 2D plane using only two kernels. We optimize rotation, scaling, position, opacity and GH coefficients. The color is rendered through alpha blending as original Gaussian splatting. Our 2DGH kernel displays great capacity for fitting an anisotropic shape like mesh triangle and also shares a merit of clear edge cut while the original 2DGS kernel causes dim and irregular edges.

\subsection{Gaussian-like Activation}
To ensure numerical stability and physical significance, we need to restrict Eq. (\ref{eq:gh_kernel}) to the range $[0,1]$. Thus we apply a new activation, Gaussian-like (GL) activation, to the Eq. (\ref{eq:gh_kernel}):
\begin{equation}
    \begin{split}
    GL(x) &= 1 - \exp(-\sigma x^{2}),
      \label{eq:gl_ac}
  \end{split}
\end{equation}
where $\sigma$ is a hyperparameter.
Sigmoid-like functions were deliberately avoided due to their observed performance degradation in our tests. We hypothesize that GL activation's superiority stems from three inherent properties:
\begin{itemize}
    \item Sigmoid activations are more prone to saturation near zero compared to GL functions.
    \item Symmetry preservation: The even symmetry of GL function aligns with the orthogonality structure of Hermite polynomials under the weight function $e^{-x^{2}}$. This compatibility maintains balanced gradient dynamics for positive/negative coefficient optimization.
    \item Spectral coherence: The exponential term $e^{-\sigma x^{2}}$ structurally resembles the weight function governing Hermite polynomial orthogonality. This similarity implicitly enhances low-frequency component representation, stabilizing coefficient optimization.
\end{itemize}

\noindent
In our experiments we empirically fix the hyperparameter $\sigma$ and set $\sigma=5$.

\subsection{Unified Initialization}

We empirically observed that the Gaussian primitive quantity has a critical impact on the result quality. The more Gaussian primitives are used, the better high-frequency details can be represented, leading to improved quantitative metrics such as PSNR, SSIM, and LPIPS.
To ensure fair comparison, we implemented a unified initialization protocol using the original 2DGS framework for all methods. This standardization guarantees identical initial Gaussian counts across different methodologies. Therefore, we can rigorously evaluate and compare the intrinsic fitting capacities of different methods.

Specifically, we train for 15000 iterations using the original 2DGS and then fix the Gaussian number. Then, we expand the Gaussian kernel with potentially higher flexibility to complete the remaining 15000-step training. In our experiments, this protocol applies to the original 2DGS, 2DGES, and our 2DGH approach.

\subsection{Floater Elimination}

\begin{figure}[t!]
\centering
\large
\includegraphics[width=\linewidth]{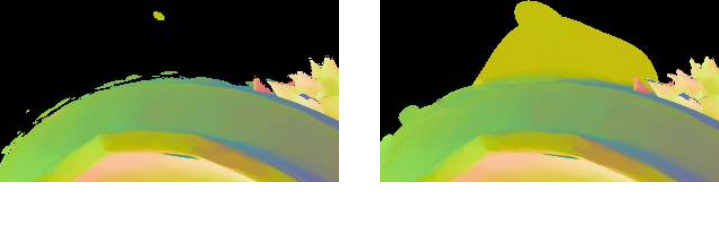}
\vspace{-1.5 em}
\caption{
\textbf{Floaters.}
In (a), we use 2DGH with a fixed learning rate for the Hermite coefficients to reconstruct the bonsai scene from the Detail dataset.
In (b), we load the final checkpoint of (a) and set all high-order GH coefficients to zero. It allows us to observe the actual positions of the primitives. This reveals that some primitives drift away from the bonsai's surface.
}  
\label{fig:no_lr_ctrl}
\end{figure}

While our method performs well in unbounded scenes, floaters occasionally emerge near sharp depth discontinuities (e.g., isolated foreground objects; see Fig.~\ref{fig:no_lr_ctrl}).
Sometimes, primitives may drift away from the object surfaces during optimization. However, the reduced position learning rate hinders the proper recovery of these displaced primitives, while the powerful fitting capability of Gaussian-Hermite (GH) kernels compensates for these artifacts in a suboptimal manner.

To address this issue, we adopt four complementary strategies:
(1) random background color supervision,
(2) enlarged position learning rate,
(3) learning rate decay for GH coefficients, and
(4) a coarse-to-fine optimization of Hermite coefficients.
All bounded-scene datasets employ random background and the coarse-to-fine strategy.
The Detail dataset, featuring the most complex geometry, utilizes all four techniques.
For DTU, we apply the enlarged position learning rate, while NeRF-Synthetic uses GH learning rate decay control.
Together, these strategies effectively suppress floater artifacts and enhance reconstruction stability.

\section{Experiments and Results}
\label{sec:expr}

\subsection{Datasets and Metrics}

We evaluate the performance of our method on both synthetic and real datasets, including Synthetic NeRF dataset~\cite{mildenhall2020nerf}, DTU dataset~\cite{jensen2014large}, Mip-NeRF 360 dataset~\cite{barron2022mip360} and a proposed synthetic Detail dataset as shown in Tab. \ref{tab:dataset_eval}.

Following the protocol of 2D Gaussian Splatting (2DGS)~\cite{huang20242dgs}, we evaluate geometric quality on Synthetic NeRF dataset, Detail dataset and DTU dataset, and assess NVS performance on Synthetic NeRF dataset, Detail dataset and Mip-NeRF 360 dataset.

For novel view synthesis and train-view rendering quality, we compare SSIM~\cite{wang2004ssim}, PSNR and LPIPS~\cite{zhang2018perceptual} among scenes on Synthetic NeRF dataset, Detail dataset and Mip-NeRF 360 dataset.

For geometry quality, we compute bidirectional Chamfer Distance (CD) between ground truth and TSDF-extracted mesh on Synthetic NeRF dataset, Detail dataset and DTU dataset. 

For training and evaluation, we keep the resolution of Synthetic NeRF dataset and Detail dataset. We follow 2DGS~\cite{huang20242dgs} and downsample the resolution of DTU dataset and Mip-NeRF 360 dataset to accelerate the process.

\begin{table}[t]
\centering
\caption{Dataset evaluation summary.}
\begin{tabular}{lccc}
\toprule
Dataset & Real/Syn & Rendering Eval & Geometry Eval \\
\midrule
Mip-NeRF 360 & Real & $\checkmark$ & -- \\
DTU & Real & -- & $\checkmark$ \\
NeRF Synthetic & Syn & $\checkmark$ & $\checkmark$ \\
Detail (Ours) & Syn & $\checkmark$ & $\checkmark$ \\
\bottomrule
\end{tabular}
\label{tab:dataset_eval}
\end{table}

\subsection{Implementation and Baseline}
We implement our 2DGH kernel by modifying the original 2DGS CUDA kernel. We add Hermite polynomials to the Gaussian function and apply a new Gaussian-like activation function. Our method is applicable to a wide range of systems employing Gaussian Splatting.

We select the original 2DGS and GES as baselines for comparison. However, the official GES implementation~\cite{hamdi2024ges} is built upon 3DGS, by which geometry extraction is challenging. Based on the GEF formulation in the GES paper, we successfully adapt it to 2DGS, resulting in a corresponding version called 2DGES, which serves as an additional baseline.

For all methods, inspired by MipNeRF360~\cite{barron2022mip360}, we use random background colors to provide supervision for transparent backgrounds. We have found that this strategy is helpful in suppressing floaters on both the Synthetic NeRF dataset and the Detail dataset. For each scene in each dataset, we run Adam for 30,000 iterations in total. The loss function follows the same form as 2DGS.

\begin{equation}
\mathcal{L} = \mathcal{L}_c + \lambda_{depth} \mathcal{L}_d + \lambda_{normal} \mathcal{L}_n
\label{eq:loss}
\end{equation}

where $\mathcal{L}_c$ denotes the RGB reconstruction loss, and $\mathcal{L}_d$ and $\mathcal{L}_n$ represent depth and normal regularization terms, respectively. We largely keep the evaluation script settings from 2DGS: the weight of the normal loss $\lambda_\text{normal}$ is 0.05 for all datasets, and the weight of the depth distortion loss $\lambda_\text{depth}$ is 1000 for DTU and 100 for the other datasets. All experiments are
conducted on an RTX 4090 GPU.

\subsection{Results and Comparison}

\begin{figure}[h]
\centering
\includegraphics[width=\linewidth]{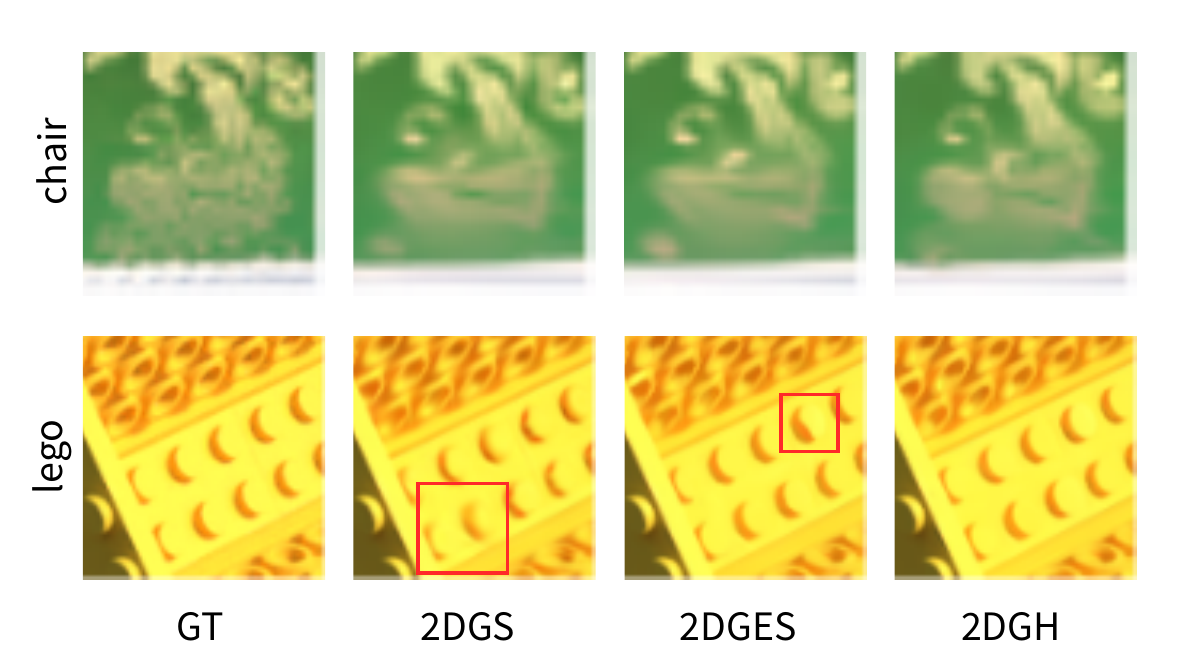} 
\vspace{-0.5 em}
\caption{
\textbf{Visual comparison of NVS on Synthetic NeRF dataset.}
In the \textit{chair} scene, 2DGH produces more natural and faithful reconstructions of the curved patterns. 
In the \textit{lego} scene, 2DGH better captures the high-frequency protrusions, demonstrating improved modeling of fine geometric details.}
\label{fig:ns_nvs}
\end{figure}

\begin{figure*}[th]
\centering
\vspace{-1.0 em}
\includegraphics[width=\linewidth]{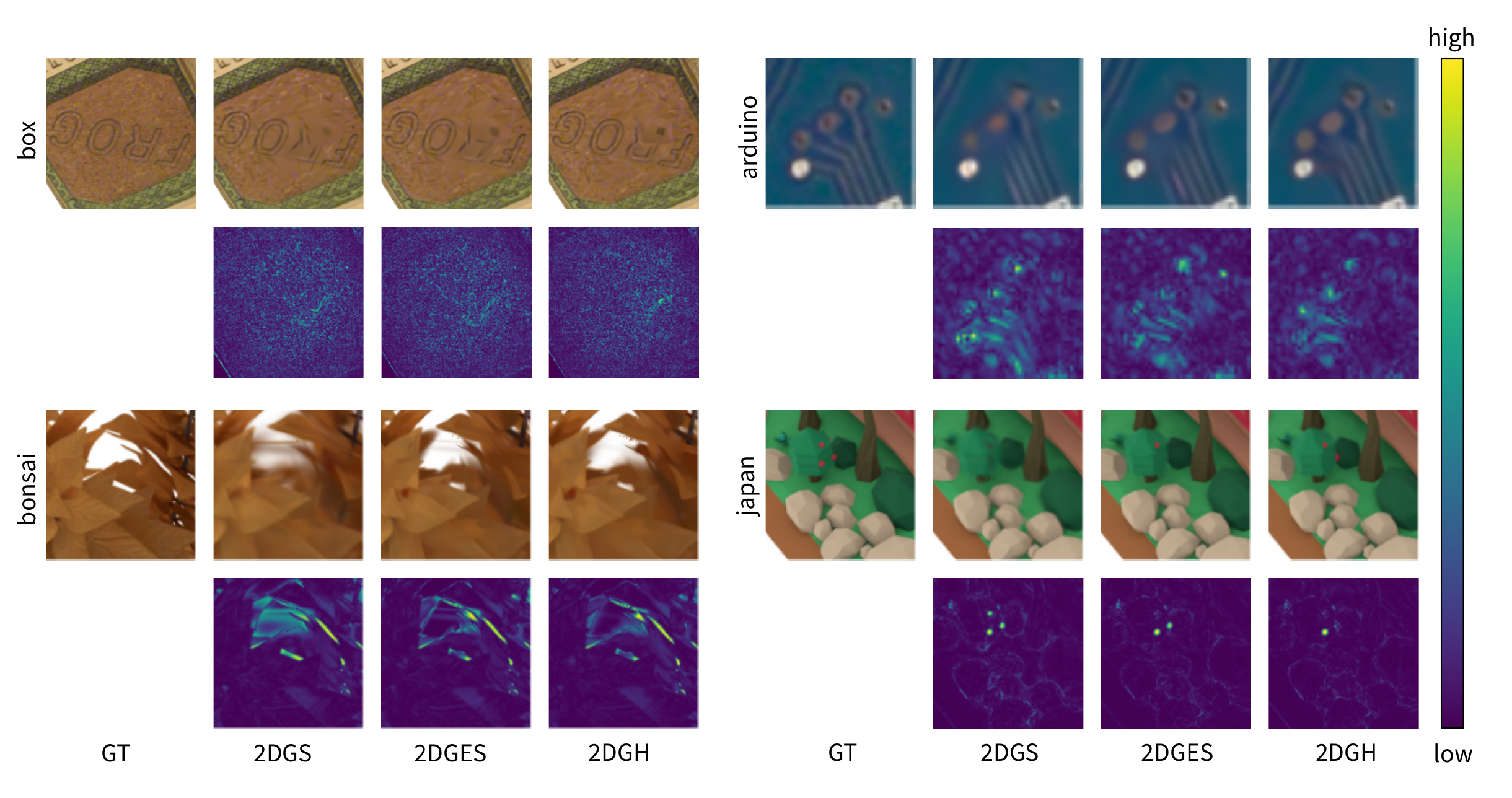} 
\vspace{-1.5 em}
\caption{
\textbf{Visual comparison of NVS on Detail dataset.}
The odd-numbered rows display the RGB rendering results, while the even-numbered rows show the differences from the ground truth. 2DGH offers the most accurate reconstruction of the fine structure and most effectively captures the slender leaves when viewed from the side.}
\label{fig:detail_nvs}
\end{figure*}

\begin{figure*}[th]
\centering
\vspace{-0.5 em}
\includegraphics[width=\linewidth]{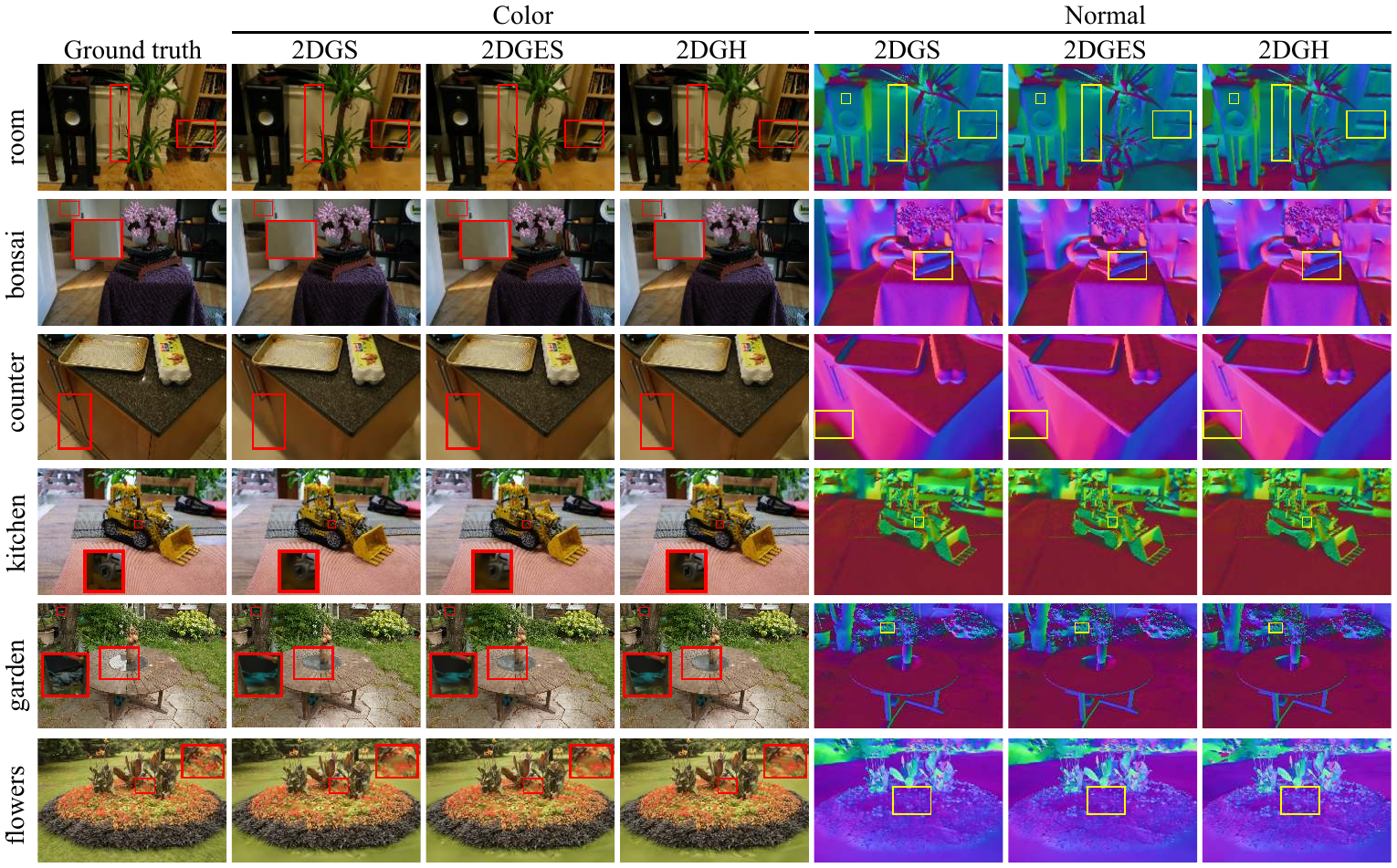}
\vspace{-1.0 em}
\caption{
\textbf{Comparison of NVS and normal results on Mip-NeRF 360 dataset.}
The left part shows the results of the RGB images, while the right part presents the results of the normals. Compared to other methods, our approach effectively captures fine and sharp discontinuous edges.
}  
\label{fig:360_nvs_full}
\end{figure*}

\hspace*{\parindent}\textbf{Train-view Rendering Quality \& Novel-view Synthesis.}
2DGS can still represent 3D scenes as radiance fields and enabling high-quality rendering and novel view synthesis. We compare novel view synthesis and train-view rendering quality of our 2DGH kernel with original 2DGS kernel on Synthetic NeRF dataset, Detail dataset and Mip-NeRF 360 dataset as shown in Tab. \ref{tab:ns_nvs}, \ref{tab:detail_nvs}, \ref{tab:360_nvs} and \ref{tab:360_train}. And we also provide Fig. \ref{fig:ns_nvs}, \ref{fig:detail_nvs} and \ref{fig:360_nvs_full} for qualitative comparison.

\begin{table}[htbp]
  \centering
  \small
  \caption{\textbf{Quantitative comparison of train-view rendering and novel-view synthesis on Synthetic NeRF dataset.} 
  We achieved the best visual results on both the training and test sets.}
  \resizebox{\linewidth}{!}{\begin{tabular}{c|ccc|ccc}
    \toprule
     &\multicolumn{3}{c|}{Novel-view} & \multicolumn{3}{c}{Train-view} \\
    Method & SSIM$\uparrow$ & PSNR (dB)$\uparrow$ & LPIPS$\downarrow$ & SSIM$\uparrow$ & PSNR (dB)$\uparrow$ & LPIPS$\downarrow$ \\
    \midrule
    2DGS & \tbest0.9673 & \tbest32.7050 & \tbest0.0354 & \tbest0.9761 & \tbest34.8842 & \tbest0.0312\\ 
    2DGES & \sbest0.9678 & \sbest32.8558 & \sbest0.0339 & \sbest0.9771 & \sbest35.2814 & \sbest0.0295\\
    2DGH & \best0.9682 & \best32.9281 & \best0.0330  & \best0.9783 & \best35.6774 & \best0.0280\\
    \bottomrule
  \end{tabular}}
  \label{tab:ns_nvs}
\end{table}

\begin{table}[htbp]
  \centering
  \small
  \caption{\textbf{Quantitative comparison of train-view rendering and novel-view synthesis on Detail dataset.}
  We achieved the best visual results on both the training and test sets.}
  \resizebox{\linewidth}{!}{\begin{tabular}{c|ccc|ccc}
    \toprule
     &\multicolumn{3}{c|}{Novel-view} & \multicolumn{3}{c}{Train-view} \\
    Method & SSIM$\uparrow$ & PSNR (dB)$\uparrow$ & LPIPS$\downarrow$ & SSIM$\uparrow$ & PSNR (dB)$\uparrow$ & LPIPS$\downarrow$ \\
    \midrule
    2DGS & \tbest0.9598 & \tbest35.6287 & \tbest0.0710 & \tbest0.9672 & \tbest37.1488 & \tbest0.0641\\ 
    2DGES & \sbest0.9630 & \sbest36.4181 & \sbest0.0632 & \sbest0.9707 & \sbest38.2613 & \sbest0.0561\\
    2DGH & \best0.9652 & \best36.6202 & \best0.0605  & \best0.9725 & \best38.3905 & \best0.0537\\
    \bottomrule
  \end{tabular}}
  \label{tab:detail_nvs}
\end{table}

\begin{table}[htbp]
  \centering
  \small
  \caption{\textbf{
Quantitative comparison of NVS on Mip-NeRF 360 dataset.}
We ensure that the number of primitives is kept the same across different methods for the same case.}
  \resizebox{\linewidth}{!}{\begin{tabular}{c|ccc|ccc}
    \toprule
    &\multicolumn{3}{c|}{Outdoor scene} & \multicolumn{3}{c}{Indoor scene} \\
    Method & SSIM$\uparrow$ & PSNR (dB)$\uparrow$ & LPIPS$\downarrow$ & SSIM$\uparrow$ & PSNR (dB)$\uparrow$ & LPIPS$\downarrow$ \\
    \midrule
    2DGS & \tbest0.7246 & \best24.961 & \tbest0.264 & \tbest0.9102 & \tbest30.017 & \tbest0.214 \\
    2DGES & \sbest0.7251 & \sbest24.561 & \sbest0.262 & \sbest0.9110 & \sbest30.050 & \sbest0.211 \\
    2DGH & \best0.7270 & \tbest24.532 & \best0.257 & \best0.9134 & \best30.127 & \best0.206 \\
    \bottomrule
  \end{tabular}}
  \label{tab:360_nvs}
\end{table}

\begin{table}[htbp]
  \centering
  \small
\caption{\textbf{
Quantitative comparison of train-view result on Mip-NeRF 360 dataset.}
We ensure that the number of primitives is kept the same across different methods for the same case.}
  \resizebox{\linewidth}{!}{\begin{tabular}{c|ccc|ccc}
    \toprule
    &\multicolumn{3}{c|}{Outdoor scene} & \multicolumn{3}{c}{Indoor scene} \\
    Method & SSIM$\uparrow$ & PSNR (dB)$\uparrow$ & LPIPS$\downarrow$ & SSIM$\uparrow$ & PSNR (dB)$\uparrow$ & LPIPS$\downarrow$ \\
    \midrule
    2DGS & \tbest0.8061 & \tbest25.7892 & \tbest0.221 & \tbest0.9222 & \tbest31.216 & \tbest0.204 \\
    2DGES & \sbest0.8104 & \sbest25.8860 & \sbest0.218 & \sbest0.9233 & \sbest31.290 & \sbest0.201 \\
    2DGH & \best0.8215 & \best26.1685 & \best0.208 & \best0.9262 & \best31.460 & \best0.195 \\
    \bottomrule
  \end{tabular}}
  \label{tab:360_train}
\end{table}

Quantitatively, our method outperforms both GES and the original 2DGS across all metrics in novel view synthesis tasks, particularly on the Detail and Synthetic NeRF datasets, as well as in indoor scenes from the Mip-NeRF 360 dataset. We hypothesize that the observed discrepancy in PSNR performance for outdoor scenes in Mip-NeRF 360 dataset, when compared to SSIM and LPIPS, stems from slight misalignment in camera poses and the fact that the local optima for PSNR do not align with those for SSIM or LPIPS.

The enhanced rendering detail can be attributed to 2DGH's ability to capture high-frequency information via high-rank Hermite polynomials. This advantage is particularly evident in our qualitative results, where our method produces sharper edges than both the original 2DGS and 2DGES. As expected, our 2DGH kernel also delivers superior performance in rendering quality evaluation for training-set views.

\textbf{Geometry Features.}
As shown in Tab. \ref{tab:ns_cd}, our method achieves a lower Chamfer Distance (CD) than both the original 2DGS and 2DGES on the Synthetic NeRF dataset. To further demonstrate our method’s capability in modeling complex geometries, we conduct additional evaluations on our custom Detail dataset, where the surface normals of ground-truth geometries are not perturbed by normal maps. The results confirm consistent and noticeable improvements in geometric accuracy.

\begin{figure}[t!]
\centering
\includegraphics[width=\linewidth]{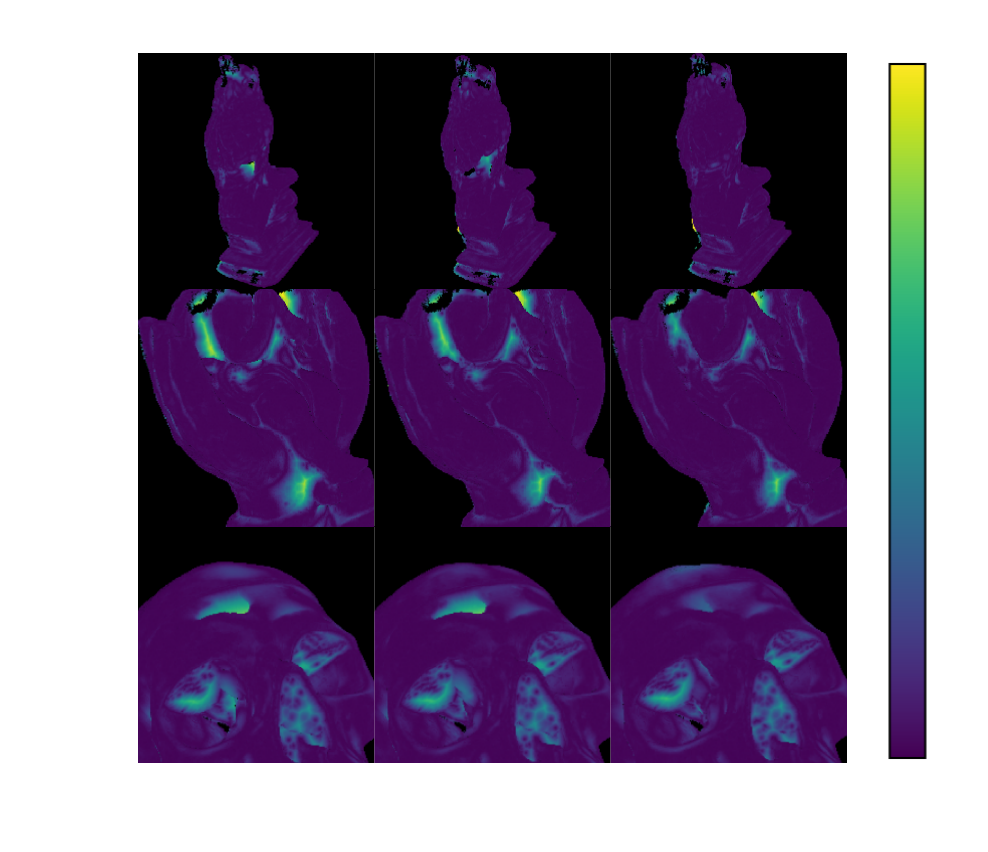}
\vspace{-1.5 em}
\caption{
\textbf{Geometry Distance.}
We use the evaluation and visualization code from 2DGS to compute the geometric errors of different methods on the DTU dataset. A viridis colorbar is used to visualize the results. Our method, 2DGH, shows lower errors, particularly in regions with rapid geometric variations.
}  
\label{fig:dtu_cd_img}
\end{figure}

For the DTU dataset, our method demonstrates significant geometric improvements, as quantitatively validated in Tab. \ref{tab:dtu_cd} and qualitatively illustrated in Fig. \ref{fig:dtu_cd_img}. The results show that our approach achieves the best Chamfer Distance (CD) in the majority of scenes.

In cases where our method does not attain the absolute best performance, it still maintains near-optimal CD values (within 0.75\% of the top results), while exhibiting substantial enhancements (up to 10\% CD reduction) in geometrically complex scenarios with intricate surface details. This evidence conclusively establishes our method's superior capability in high-fidelity geometric reconstruction, particularly when handling challenging topological structures.

\begin{table*}[htbp]
  \centering
  \footnotesize
    \caption{\textbf{Quantitative comparison of geometric reconstruction on DTU dataset.} Our 2DGH method obtains the lowest mean Chamfer Distance (0.7962), demonstrating consistent improvements over 2DGS (0.8230), 2DGES (0.8192), and GOF (0.8037). We further present the number of Gaussians used in 2DGX (2DGS, 2DGES, and 2DGH) versus GOF\cite{yu2024gaussian}. To emphasize the representational capacity of Gaussians, we adjust the number of Gaussians in GOF to ensure the comparison is as fair as possible. }
  \resizebox{\linewidth}{!}{\begin{tabular}{c|cccccccccccccccc}
    \toprule
    Method & 24 & 37 & 40 & 55 & 63 & 65 & 69 & 83 & 97 & 105 & 106 & 110 & 114 & 118 & 122 & Mean \\
    \midrule
    \multicolumn{17}{c}{Chamfer Distance} \\
    \midrule
    2DGS & \tbest{0.5132} & \sbest{0.9395} & \tbest{0.4540} & 0.4268 & \sbest{0.9074} & 0.8583 & 0.8540 & \sbest{1.4004} & \tbest{1.2525} & \sbest{0.8222} & 0.7333 & \sbest{1.3967} & \sbest{0.4076} & 0.7629 & \tbest{0.6157} & 0.8230 \\
    2DGES & 0.5155 & 0.9449 & 0.4621 & \tbest{0.4265} & \tbest{0.9095} & \sbest{0.8483} & \tbest{0.8234} & 1.4147 & \sbest{1.2393} & \tbest{0.8266} & \tbest{0.7059} & \tbest{1.4021} & \best{0.4039} & \tbest{0.7206} & 0.6445 & \tbest{0.8192} \\
    2DGH & \best{0.4905} & \tbest{0.9400} & \sbest{0.4265} & \sbest{0.4144} & \best{0.8924} & \best{0.8197} & \best{0.7966} & \tbest{1.4103} & \best{1.1774} & 0.8271 & \sbest{0.6957} & 1.4024 & \tbest{0.4078} & \sbest{0.6875} & \sbest{0.5554} & \best{0.7962} \\
    \midrule
    GOF & \sbest{0.5033} & \best{0.8501} & \best{0.3952} & \best{0.3727} & 1.2936 & \tbest{0.8522} & \sbest{0.7988} & \best{1.2637} & 1.3822 & \best{0.6833} & \best{0.6614} & \best{1.3527} & 0.4567 & \best{0.6652} & \best{0.5251} & \sbest{0.8037} \\
    \midrule
    \multicolumn{17}{c}{Number of Gaussians} \\
    \midrule
    2DGX & 293717 & 363458 & 652078 & 304693 & 175308 & 175916 & 178835 & 144963 & 227951 & 156338 & 118462 & 108579 & 130341 & 108994 & 115795 & /\\
    GOF & 297908 & 368082 & 681765 & 308654 & 188363 & 207234 & 180697 & 161812 & 274676 & 168525 & 119416 & 114845 & 130408 & 109177 & 127615 & /\\
    \bottomrule
  \end{tabular}}
  \label{tab:dtu_cd}
\end{table*}

Apart from CD metrics, we particularly emphasize the normal enhancement capabilities of our 2DGH method. As evidenced in Fig. \ref{fig:360_nvs_full}, our 2DGH method generates notably sharper normal boundaries compared to baseline approaches. Our analysis reveals that conventional Gaussian-shaped elliptical kernels are ill-suited for modeling high-frequency variations at geometric boundaries. This inherent limitation persists even when the GES variant introduces additional high-frequency terms, as it fails to fully resolve boundary irregularities. Only through 2DGH's enhanced anisotropic deformation capability---achieved via high-order Hermite basis functions---can this fundamental limitation be systematically addressed.
\begin{table}[htbp]
  \centering
  \footnotesize
    \caption{\textbf{Quantitative comparison of Chamfer Distance$\downarrow$ ($\times 10^{-3}$) on Synthetic NeRF dataset.}
    Compared to 2DGS and 2DGES, our method achieves superior average performance. These quantitative results are generally consistent with the observations from the qualitative experiments.}
  \begin{tabular}{c|cc}
    \toprule
    Method & Synthetic NeRF Mean CD$\downarrow$ & Detail Mean CD$\downarrow$ \\
    \midrule
    2DGS & \tbest2.955 & \tbest 2.110 \\
    2DGES & \sbest2.895 & \sbest 2.011  \\
    2DGH & \best2.864 & \best1.934 \\
    \bottomrule
  \end{tabular}
  \label{tab:ns_cd}
\end{table}

\textbf{Overhead.}
In the previous experiments, a fair comparison is ensured by using an equal number of primitives across all methods. Given that our method involves more parameters, it naturally incurs a slightly higher memory overhead (around 10\%) and a longer training duration (around 10-20\%). Table~\ref{tab:memory_time_compact} quantifies the memory overhead and training durations under different configurations.
Since our method explicitly focuses on recovering both material and geometry, we consider the slightly longer training time acceptable.
More importantly, owing to the superior expressive power of the Gaussian–Hermite representation, our approach can reduce the number of primitives while still maintaining stronger performance in both rendering quality and geometric reconstruction. Tab. \ref{tab:ns_nonequal_gaussians} illustrates that with less memory usage, 2DGH can achieve higher novel view synthesis quality and more accurate shape than original 2DGS and is a competitive method compared with GES.

\begin{table}[htbp]
  \centering
  \small
    \caption{
  \textbf{Quantitative comparison on Synthetic NeRF dataset with unequal numbers of primitives.}
  NUM is referred to the number of primitives. Our method demonstrates better novel-view synthesis and geometric results with less memory overhead and less primitives compared to 2DGS.
  }
  \resizebox{\linewidth}{!}{\begin{tabular}{c|ccccccc}
    \toprule
    Method & SSIM$\uparrow$ & PSNR (dB)$\uparrow$ & LPIPS$\downarrow$ & CD$\downarrow$ & NUM$\downarrow$ & Memory\,(MB)$\downarrow$\\
    \midrule
    2DGS & \tbest0.9673 & \tbest32.705 & \tbest0.0354 & \tbest0.002955 &\tbest73118 & \tbest18.585\\ 
    2DGES & \sbest0.9677 & \sbest32.846 & \sbest0.0341 & \sbest0.002887 &\sbest69190 &\best18.291\\
    2DGH & \best0.9677 & \best32.851 & \best0.0340 & \best0.002786 &\best61654 &\sbest18.525\\
    \bottomrule
  \end{tabular}}
  \label{tab:ns_nonequal_gaussians}
\end{table}

\begin{table}[htbp]
  \centering
  \scriptsize
  \caption{\textbf{Quantitative comparison of the average storage (MB) and training durations(min) across different methods in the previous rendering quality, NVS and geometric reconstruction experiments.} Each cell shows \textit{Mem (Time)}. To ensure fairness, we set the number of primitives to be the same for each case. }
  \begin{tabular}{c|cccc}
    \toprule
    Dataset & Synthetic NeRF & Detail & DTU & Mip-NeRF 360 \\
    \midrule
    2DGS  & 18.6 (4.6) & 10.4 (5.4) & 50.5 (8.2) & 451.2 (21.9) \\
    2DGES & 18.9 (4.7) & 10.6 (5.3) & 51.3 (8.2) & 458.6 (22.3) \\
    2DGH  & 21.3 (5.1) & 12.0 (5.7) & 58.0 (9.2) & 517.7 (25.3) \\
    \bottomrule
  \end{tabular}
  \label{tab:memory_time_compact}
\end{table}


We demonstrate that there exists additional design space for trading off between model parameters and the number of primitives. Our approach provides a highly flexible strategy to achieve it through adjustment of the maximum rank in Gaussian-Hermite coefficients. Compared to 2DGS, our proposed 2DGH method can reduce the number of primitives by over 15\%. This advantage may become particularly beneficial for large-scale datasets, where the number of Gaussians often becomes an optimization bottleneck.

\subsection{Ablation Study}

\hspace*{\parindent}\textbf{Gaussian-Hermite Polynomial Rank.}
We investigate how the rank of Gaussian-Hermite (GH) coefficients affects rendering quality and geometry reconstruction metrics. In our formulation, the coefficients $c_{mn}$ are optimizable GH parameters, where $N$ and $M$ represent the maximum rank numbers in orthogonal directions. Tab. \ref{tab:ab2} compares the performance across different $M,N$ configurations.

\begin{table}[htbp]
  \centering
  \footnotesize
    \caption{
  \textbf{Quantitative ablation study on the amount of optimizable GH coefficients.} As the highest order of Hermite polynomials increases, 2DGH exhibits stronger representational capacities, as evidenced by improvements in both rendering and geometric metrics.
  }
  \resizebox{\linewidth}{!}{\begin{tabular}{c|ccccc}
    \toprule
    Setting & SSIM$\uparrow$ & PSNR (dB)$\uparrow$ & LPIPS$\downarrow$ & mean CD$\downarrow$\\
    \midrule
     M=N=0(2DGS + GL activation) & 0.9675 & 32.7761 & 0.0343 & 0.002982\\
    M=N=1  & \tbest0.9677 & \tbest32.8708 & \tbest0.0339 & \tbest0.002933\\
    M=N=2 & \sbest0.9680 & \best32.9325 & \sbest0.0336 & \best0.002863\\
    M=N=3 & \best0.9682  & \sbest32.9281 & \best0.0330 & \sbest0.002864\\
    \bottomrule
  \end{tabular}}
  \label{tab:ab2}
\end{table}

As expected, increasing M and N improves the performance of 2DGH, though with diminishing returns—the gains in reconstruction metrics gradually saturate as the rank grows.

\begin{figure}[t!]
\centering
\includegraphics[width=\linewidth]{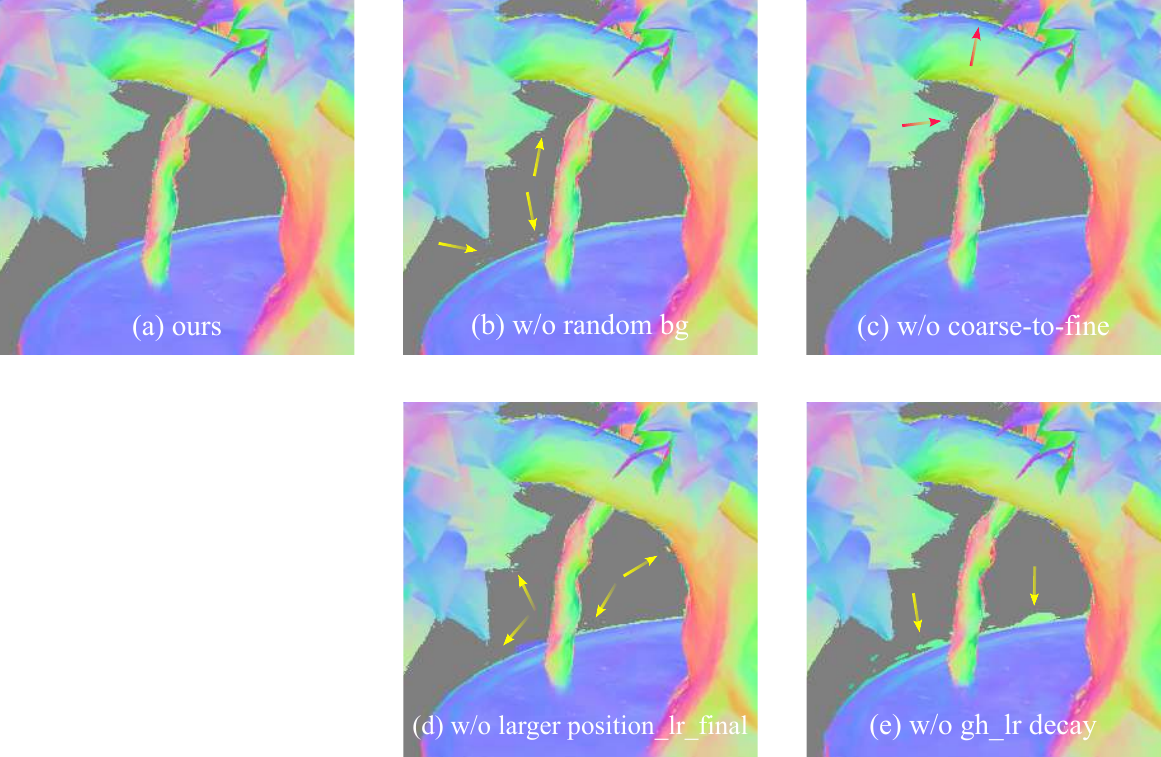}
\vspace{-1.5em}
\caption{
\textbf{Ablation study on floater elimination.}
\maybeColor{Visualization of surface normals from the Bonsai scene in the Detail dataset, which reveals the spatial distribution of Gaussian opacity.}
(a) Ours (full strategy) successfully suppresses floaters. 
(b) w/o random\_bg: the absence of random background supervision results in floaters in free space, as indicated by the arrows. 
(c) w/o coarse-to-fine: introducing all GH orders at once produces slightly more floaters compared to the full strategy. 
(d) w/o larger position\_lr\_final: Increasing this final value raises the overall positional learning rate. 
(e) w/o gh\_lr decay: using a large fixed GH coefficient learning rate leads to the most noticeable floater artifacts.
}
\label{fig:ablation_floater}
\end{figure}

\textbf{Floater Elimination.}
To validate the effectiveness of our proposed strategies in suppressing floater artifacts, we conducted a series of ablation experiments, as shown in Figure~\ref{fig:ablation_floater}:
\begin{itemize}
    \item \textbf{w/o random\_bg}: We remove random background supervision and instead use a fixed white background, which leads to noticeable floater artifacts.
    \item \textbf{w/o coarse-to-fine}: Introducing all GH coefficients at the beginning has a relatively milder impact on floaters compared to other factors, but still makes the positional optimization of Gaussians more prone to getting stuck in local minima.
    \item \textbf{w/o larger position\_lr\_final}: Retaining the original 2DGS position learning rate schedule prevents drifted primitives from being pulled back to the surface, while GH coefficients attempt to compensate in a suboptimal manner.
    \item \textbf{w/o gh\_lr decay}: Without a decayed learning rate for GH coefficients, Hermite terms tend to overfit local structures and introduce additional floaters. A large fixed GH learning rate may also destabilize the optimization.
\end{itemize}

We also report quantitative results for NVS and geometry reconstruction in Table~\ref{tab:ab_floater}. Interestingly, the absence of individual strategies does not lead to significant changes in quantitative performance, and all settings still outperform the baseline, highlighting the superiority of our Gaussian-Hermite formulation.

\begin{table}[htbp]
  \centering
  \footnotesize
  \caption{
  \textbf{Quantitative ablation study on Floater Elimination Strategy.} 
  }
  \resizebox{\linewidth}{!}{\begin{tabular}{c|ccccc}
    \toprule
    Setting & SSIM$\uparrow$ & PSNR (dB)$\uparrow$ & LPIPS$\downarrow$ & mean CD$\downarrow$ \\
    \midrule
    Ours (full strategy) & 0.9653 & 36.7159 & 0.0604 & 0.001952 \\
    w/o random\_bg & 0.9642 & 36.7499 & 0.0619 & 0.001954 \\
    w/o coarse-to-fine & 0.9655 & 36.7303 & 0.0600 & 0.001981 \\
    w/o larger position\_lr\_final & 0.9650 & 36.7051 & 0.0608 & 0.002002 \\
    w/o gh lr decay & 0.9669 & 36.6634 & 0.0551 & 0.001933 \\
    \bottomrule
  \end{tabular}}
  \label{tab:ab_floater}
\end{table}

Overall, these results demonstrate that our full strategy---including random background supervision, enlarged position learning rate, GH learning rate decay, and the coarse-to-fine scheme---works synergistically to effectively suppress floater artifacts and improve both geometric reconstruction and rendering stability.

\subsection{Limitation}

\hspace*{\parindent}\textbf{Specular Reflection.}
While 2DGH has demonstrated promising results in 3D reconstruction, it faces challenges when handling reflective surfaces—a limitation common to many existing reconstruction methods. This difficulty primarily arises from the effects of global illumination. Potential solutions include incorporating ray tracing and physically based rendering (PBR) material decomposition \cite{ye20243d,jiang2024gaussianshader,gu2025irgs}, or leveraging neural networks to model high-frequency view-dependent appearance changes \cite{yang2024spec}. Regardless of the chosen solution, it is orthogonal to our method and can be combined with it.

\textbf{Rendering Speed.}
To compute the Hermite polynomial summation, we introduced additional computational overhead and parameter packaging. Although our method can reduce the number of Gaussians, the rendering speed bottleneck in most scenes is not solely determined by Gaussian count. As shown in Tab. \ref{tab:fps}, on the Synthetic NeRF dataset, our method experiences a 20–30\% reduction in forward rendering speed compared to 2DGS. The reduction in the number of Gaussians is insufficient to fully compensate the loss in rendering speed.

This issue might be mitigated through more efficient low-level implementations. Alternatively, pruning methods that remove Gaussians with minimal visual contribution could also help \cite{fan2024lightgaussian, yuan20251000+, niemeyer2024radsplat}. Compared to 3DGS, the primary purpose of 2DGS is to enable more accurate reconstruction of digital assets, including geometric meshes, which are better suited for downstream tasks such as visual preview, editing, and processing. In particular, for visual preview, extracting meshes and materials from 2DGS or our 2DGH could provide a more effective way to achieve higher frame rates.

\begin{table}[h]
\centering
    \footnotesize
\caption{\textbf{Comparison of mean forward rendering speed (FPS) and mean number of Gaussians under unequal and equal Gaussian counts on the Synthetic NeRF dataset}}
\begin{tabular}{l|cc|cc}
\toprule
\textbf{Method} & \multicolumn{2}{c|}{\textbf{Unequal Gaussian Count}} & \multicolumn{2}{c}{\textbf{Equal Gaussian Count}} \\
                & FPS $\uparrow$ & NUM $\downarrow$ & FPS $\uparrow$ & NUM \\
\midrule
2DGS           & 585.62 & 79863.25 & 585.62 & 79863.25 \\
2DGES          & 526.70 & 77330.00 & 497.57 & 79863.25 \\
2DGH           & 437.73 & 69367.13 & 407.40 & 79863.25 \\
\bottomrule
\end{tabular}
\label{tab:fps}
\end{table}



\section{Discussion and Conclusion}
\label{sec:conclusion}

Through this research, we have introduced a family of Gaussian-Hermite kernels as a unified mathematical representation in Gaussian Splatting, enabling the creation of sharp boundaries within the Gaussian representation. We have incorporated arbitrary deformation capabilities and significantly enhanced anisotropy in the Gaussian representation, reducing the gap between Gaussian primitives and mesh facets. 

To handle high-order coefficients, we have developed a novel activation function, while preserving the symmetry and physical meaning of the Hermite coefficients. This also challenges the traditional view that the primitives used for splatting must be Gaussian-based functions.

The 2DGS framework allows arbitrary modification to the kernel function within the local coordinate system. This raises the question of whether there is a complete set of bases that could potentially offer largest improvements over all alternative kernels. One intriguing possibility is the use of Fourier-based functions, which may provide enhanced representational capabilities. We consider this to be a subject that warrants comprehensive investigation in the future. 

Our method exhibits excellent generalization and can serve as an add-on feature that can be seamlessly integrated into any 2DGS-based or planar-based method\cite{chen2024pgsr}, as well as various downstream tasks.
In principle, 3DGS and any pipeline built upon it can also adopt our approach. However, since the projection mechanism in 3DGS is more complex than that in 2DGS, it would require a redesign of the projection transformation. Considering that our focus lies in enhancing the expressive capacity of individual Gaussian primitives, such an extension falls beyond the scope of this work.
Nevertheless, it is worth noting that the limited representational capability observed in 3DGS is fundamentally similar to that of 2DGS.

Overall, the proposed Gaussian-Hermite kernel-based approach has demonstrated remarkable performance in both geometry reconstruction and novel-view synthesis tasks, outperforming traditional Gaussian Splatting kernel. It is also conceptually intriguing to reflect on the analogy between atomic orbitals and Gaussian primitives, where Gaussian–Hermite polynomials provide an elegant bridge between physics and computer graphics.

\section*{Acknowledgments}
This work was supported by the National Key R\&D Program of China (2023YFC3305600). This work was also supported by THUIBCS, Tsinghua University, and BLBCI, Beijing Municipal Education Commission. Feng Xu is the corresponding author.

\ifCLASSOPTIONcaptionsoff
  \newpage
\fi



\bibliographystyle{IEEEtran}
\bibliography{IEEEabrv,aaatemplate}

\vfill


\end{document}